\documentclass[runningheads]{llncs}
\usepackage{graphicx}

\usepackage{epsfig}
\usepackage{tikz}
\usepackage{comment} 
\usepackage{amsmath,amssymb} 
\usepackage{color}
\usepackage{bm}
\usepackage{booktabs}
\usepackage{multirow}
\usepackage{subcaption}
\usepackage{makecell}
\usepackage{wrapfig}

\usepackage[ruled,vlined]{algorithm2e}
\usepackage{setspace}
\usepackage[title]{appendix}

\begin{document}
\pagestyle{headings}
\mainmatter
\def\ECCVSubNumber{2844}  

\title{Exchangeable Deep Neural Networks for \\ Set-to-Set Matching and Learning} 

\titlerunning{Deep Set-to-Set Matching and Learning}
%
\author{Yuki Saito\inst{1,2}\orcidID{0000-0003-0492-414X} \and
Takuma Nakamura\inst{1}\orcidID{0000-0001-7904-4724} \and
Hirotaka Hachiya\inst{3}\orcidID{0000-0003-3748-4101} \and \\
Kenji Fukumizu\inst{4,2}\orcidID{0000-0002-3488-2625}}
\authorrunning{Y. Saito et al.}
%
\institute{
ZOZO Research, Jingumae, Shibuya, Tokyo, Japan\\
\email{\{yuki.saito,takuma.nakamura\}@zozo.com} \and
The Graduate University for Advanced Studies, SOKENDAI, Tachikawa, Tokyo, Japan \and
Wakayama University, Wakayama-shi, Wakayama, Japan\\
\email{hhachiya@wakayama-u.ac.jp} \and
The Institute of Statistical Mathematics, Tachikawa, Tokyo, Japan \\
\email{fukumizu@ism.ac.jp}
}
\maketitle

\begin{abstract}
Matching two different sets of items, called heterogeneous set-to-set matching problem, has recently received attention as a promising problem. The difficulties are to extract features to match a correct pair of different sets and also preserve two types of exchangeability required for set-to-set matching: the pair of sets, as well as the items in each set, should be exchangeable. In this study, we propose a novel deep learning architecture to address the abovementioned difficulties and also an efficient training framework for set-to-set matching. We evaluate the methods through experiments based on two industrial applications: fashion set recommendation and group re-identification. 
In these experiments, we show that the proposed method provides significant improvements and results compared with the state-of-the-art methods, thereby validating our architecture for the heterogeneous set matching problem.

\keywords{set to set matching; deep learning; permutation invariance}
\end{abstract}

\section{Introduction}\label{sec:Intro}
Matching pairs of data is a crucial part of many machine learning tasks, including recommendation~\cite{sarwar2001item,rendle2010factorizing,ijcai2019-389}, person re-identification (re-id)~\cite{zheng2015scalable}, image search~\cite{DBLP:journals/corr/WangSLRWPCW14}, and face recognition~\cite{parkhi2015deep}, as typical industrial applications. Over the past decade, a deep learning framework for matching up data, e.g., images, has served as the core of such systems. 

\begin{figure}[t]
\centering
\includegraphics[width=0.7\textwidth]{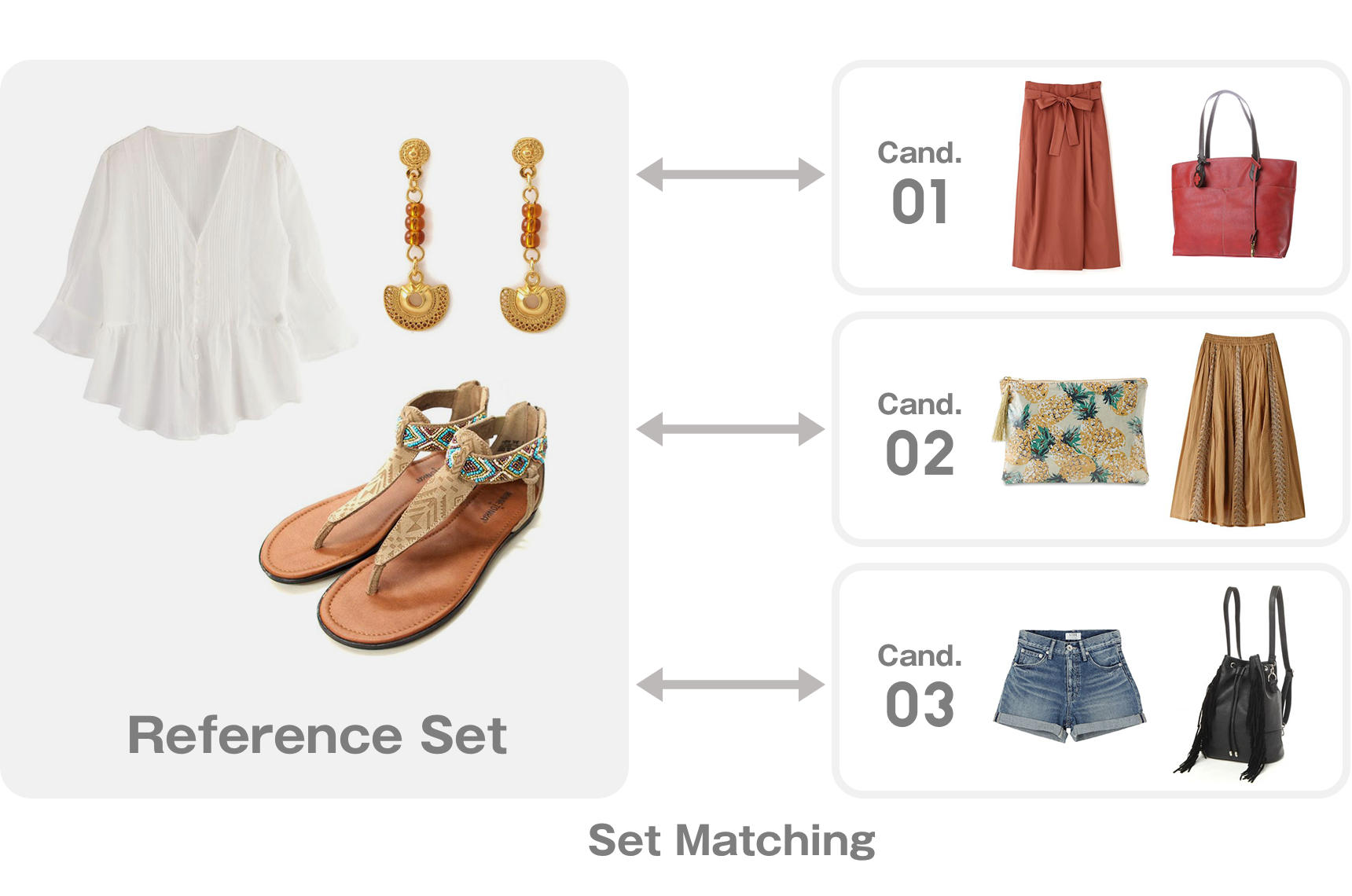} 
\caption{One of the main questions that set-to-set matching attempts to answer is as follows: which candidate is more compatible than others with the reference set? Here, we consider the matching of the reference set and the respective candidate set and then selecting the best pair. }
\label{fig:title}
\end{figure}

Aside from these tasks, set-to-set matching, which is an extension of multiple instance matching, has recently been identified as an important element in various applications required by emerging web technologies or services. 
A representative example in e-commerce is fashion recommendation, where a group of fashion items deemed to match the collection of fashion items already owned by a user is recommended. Regarding the group as an unordered set, we can consider this task a set-to-set matching problem, as shown in Figure~\ref{fig:title}. Another example is group re-identification (group re-id) in surveillance systems \cite{DBLP:journals/corr/LisantiMBF17,Xiao:2018:GRL:3240508.3240539,lin2019group}, which has recently started implementing a function to track known groups of suspicious persons or criminals, a task that can also be simplified as a set-to-set matching problem. Other examples include image-set retrieval~\cite{gao2018group,feng2017deep}, image-set classification~\cite{lu2015multi}, image-set reconstruction~\cite{liu2019exploring}, person re-id~\cite{liu2017quality}, taxonomy matching~\cite{doi:10.1111/maps.13428}, cross-lingual matching~\cite{iwata2017unsupervised}, relational data matching~\cite{iwata2015unsupervised}, and face verification~\cite{liu2019permutation,xie2018comparator}. Earlier studies have also explored face recognition as a set-to-set matching problem~\cite{shakhnarovich2002face,arandjelovic2005face,cevikalp2010face,yamaguchi1998face} and next-basket recommendation~\cite{rendle2010factorizing}.

Set matching scenarios can be grouped into two classes: homogeneous set matching and heterogeneous set matching. In the former, two positive sets comprising the same instances, such as the images of the face of the same person, are to be matched. Except for variations such as differences in illumination or pose in the images, both sets contain similar instances. This scenario has been investigated in several studies~\cite{gao2018group,lu2015multi,feng2017deep,liu2017quality,liu2019permutation,xie2018comparator,shakhnarovich2002face,arandjelovic2005face,cevikalp2010face,yamaguchi1998face,liu2019exploring}. In the heterogeneous case, the instances within paired sets can be considerably different, as is the case in fashion recommendation and group re-id.
To the best of our knowledge, there are very few studies on constructing deep learning models for heterogeneous set matching.
We consider that matching heterogeneous sets is a more difficult task and requires a strong learning architecture to match different sets.

Furthermore, another fundamental difficulty in set-to-set matching, compared with ordinary data matching, 
lies in the two types of exchangeability required: exchangeability between the pair of sets and invariance across different permutations of the items in each set. A function that calculates a matching score should provide an invariant response, regardless of the order of the two sets, or the permutations of the items.

The main focus of this paper is an architecture that preserves the aforementioned exchangeability properties, and at the same time, realizes a high performance in heterogeneous set matching tasks.
In this study, we argue that allowing the feature extractor and matching layer to include interactions between the two sets is crucial to identify matching pairs among different items. We propose a deep learning model for (1) feature extraction, named {\it cross-set feature transformation} (CSeFT), which iteratively provides the interactions between the pair of sets to each other in the intermediate layers. Our novel functions, {\it attention-} and {\it affinity-based functions}, organize the CSeFT spanning two different sets in the feature spaces, thereby improving the feature representations. The proposed architecture also includes (2) a matching layer, named {\it cross-similarity function} (CS function), that calculates the matching score between the features of the set members across the two sets accurately. Our model guarantees both types of exchangeability in the modules. Figure~\ref{fig:block} shows the proposed architecture.

We examine the set-to-set matching problem in a supervised setting, where examples of correctly paired sets are deployed as training data. The objective is to train the feature extractor and matching layer in an end-to-end manner such that the appropriate sets of features to be matched can be extracted. To train the model efficiently, we also propose a novel training framework, {\it $K$-pair-set loss}.
Following training, the model is then used to find correct pairs of sets among a group of candidates.

The effectiveness of our approach is demonstrated in two real-world applications. First, we consider fashion set matching, where provided examples of the outfits are used as correct combinations of items (clothes). Using a large number of examples of the outfits in the form of images, we aim to match the correct pair of defined sets using the IQON dataset~\cite{DBLP:journals/corr/abs-1807-03133}. Since two positive sets include images of different fashion items, we regard this case as heterogeneous set matching.
Next, we evaluate our methods through group re-id experiments using two datasets, a new extension of the Market-1501 dataset~\cite{zheng2015scalable} (Market-1501 Group) and the Road Group dataset~\cite{Xiao:2018:GRL:3240508.3240539}. 
We use the Market-1501 Group dataset to analyze sensitivity to noises or outliers in set matching and the Road Group dataset as a more practical search task. Considering group membership change caused by the noises, we regard group re-id as a heterogeneous set matching problem.
In the fashion set matching and group re-id experiments performed on the Market-1501 Group dataset, our methods show significant improvements compared with the results of baseline methods. Moreover, using the data augmentation method that we developed for the pair set (set-aug), our methods show competitive results without using any external datasets or spatial layout information on the Road Group dataset.

The main contributions of this paper are as follows. (i) A novel deep learning architecture is proposed to provide the two types of exchangeability required for set-to-set matching. (ii) The proposed feature extractors using the interactions between two sets are shown to extract better features for heterogeneous set matching. (iii) A new loss function, $K$-pair-set loss, is proposed and provide better performances in our tasks. (iv) We introduce set-input methods into group re-id tasks (Road Group) using a new set-data augmentation, thereby showing competitive results without using external datasets or spatial relations. (v) The proposed models show state-of-the-art results for the fashion set matching and group re-id, supporting the claim that the interactions and exchangeability improve the accuracy and robustness of the set-matching procedure.

\section{Preliminaries: Set-to-Set Matching} \label{sec:Preliminaries}
We introduce the necessary notation as follows.
Let $\bm{x}_n,\bm{y}_m \in \mathfrak{X} = \mathbb{R}^d$ be feature vectors representing the features of each individual item. Let $\mathcal{X}=\{\bm{x}_1,...,\bm{x}_{N}\}$ and $\mathcal{Y}=\{\bm{y}_1,...,\bm{y}_{M}\}$ be {\it sets} of these feature vectors, where $\mathcal{X},\mathcal{Y} \in 2^\mathfrak{X}$.

The function $f:2^\mathfrak{X}\times 2^\mathfrak{X}\to \mathbb{R}$ calculates a matching score between the two sets $\mathcal{X}$ and $\mathcal{Y}$. Guaranteeing the exchangeability of the set-to-set matching requires that the matching function $f(\mathcal{X},\mathcal{Y})$ is {\it symmetric} and {\it invariant} under any permutation of items within each set.

We consider tasks where the matching function $f$ is used per pair of sets~\cite{zhu2013point} to select a correct matching. Given candidate pairs of sets $(\mathcal{X},\mathcal{Y}^{(k)})$, where $\mathcal{X},\mathcal{Y}^{(k)} \in 2^\mathfrak{X}$ and $k \in \{1,\cdots,K\}$, we choose $\mathcal{Y}^{(k^*)}$ as a correct one so that $f(\mathcal{X},\mathcal{Y}^{(k^*)})$ achieves the maximum score from amongst the $K$ candidates.
In this study, a supervised learning setting is considered, where the function $f$ is trained to
classify the correct pair and unmatched pairs. 

\subsection{Mappings of Exchangeability} \label{subsec:set}

We present a brief review on several notions of exchangeability, which are used in building our models.

\noindent {\bf Permutation Invariance.}
A set-input function $f$ is said to be {\em permutation invariant} if 
\begin{equation}
  f(\mathcal{X},\mathcal{Y})=f(\pi_x \mathcal{X}, \pi_y \mathcal{Y})
\label{eq:pi2}
\end{equation}
for permutations $\pi_x$ on $\{1, \ldots, N\}$ and $\pi_y$ on  $\{1, \ldots, M\}$.

\noindent {\bf Permutation Equivariance.}
A map $f:\mathfrak{X}^N\times \mathfrak{X}^M \to \mathfrak{X}^N$ is said to be {\em permutation equivariant} if
\begin{equation}
f(\pi_x \mathcal{X}, \pi_y \mathcal{Y})=\pi_x f(\mathcal{X}, \mathcal{Y})
\label{eq:pe2}
\end{equation}
for permutations $\pi_x$ and $\pi_y$, where $\pi_x$ and $\pi_y$ are on $\{1, \ldots, N\}$ and $\{1, \ldots, M\}$, respectively. Note that $f$ is permutation invariant for permutations within $\mathcal{Y}$.

\noindent {\bf Symmetric Function.}
A map $f:2^\mathfrak{X} \times 2^\mathfrak{X} \to \mathbb{R}$ is said to be {\em symmetric} if
\begin{equation}
  f(\mathcal{X}, \mathcal{Y})=f(\mathcal{Y}, \mathcal{X}).
\label{eq:symmetric}
\end{equation}

\noindent {\bf Two-Set-Permutation Equivariance.}
Given $\mathcal{Z}^{(1)} \in \mathfrak{X}^N$ and $\mathcal{Z}^{(2)} \in \mathfrak{X}^M$,
a map $f:\mathfrak{X}^* \times \mathfrak{X}^* \to \mathfrak{X}^* \times \mathfrak{X}^*$
is said to be {\em two-set-permutation equivariant} if
\begin{equation}
  p f(\mathcal{Z}^{(1)}, \mathcal{Z}^{(2)})=f(\mathcal{Z}^{(p(1))}, \mathcal{Z}^{(p(2))})
\label{eq:pes}
\end{equation}
for any permutation operator $p$ exchanging the two sets, where $\mathfrak{X}^* = \cup_{n=0}^\infty \mathfrak{X}^n$ indicates a sequence of arbitrary length such as $\mathfrak{X}^N$ or $\mathfrak{X}^M$.

\begin{figure}[t]
\centering
\includegraphics[width=0.999\textwidth]{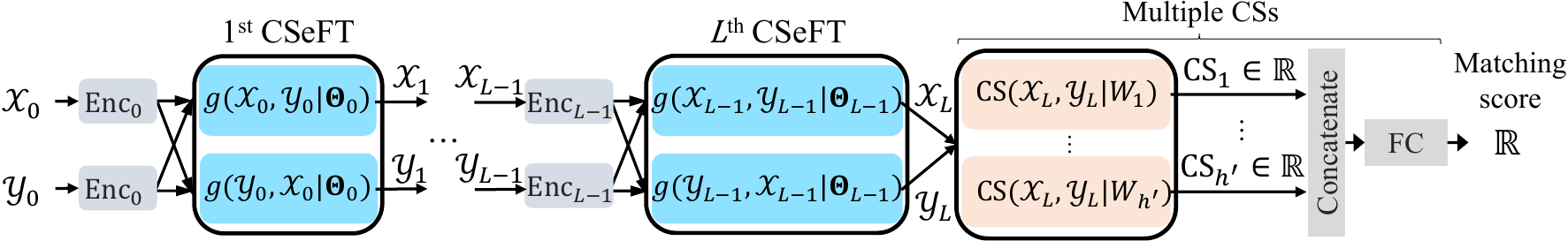} 
\caption{Our model calculates a matching score between the paired sets. $\mathrm{Enc}_i$, CSeFT, CS, and $\mathrm{FC}$ indicate an $(i+1)$-th (one-layered) encoder sharing weights within the same layer, cross-set feature transformation, cross-similarity function, and fully connected layer, respectively. We exclude the multihead structure in $g$.}
\label{fig:block}
\end{figure}

\section{Matching and Learning for Sets} \label{sec:Proposed_Method}

\subsection{Cross-Set Feature Transformation} \label{subsec:CSeFT}
We construct the architecture of the feature extractor, which transforms sets of features using the interactions between the pair of sets, and extracts the desired features to be matched in the post-processing stages (Section~\ref{subsec:as}).

\begin{figure}[t]
\centering
\begin{minipage}{.4\textwidth}
  \centering
  \includegraphics[scale = 0.38]{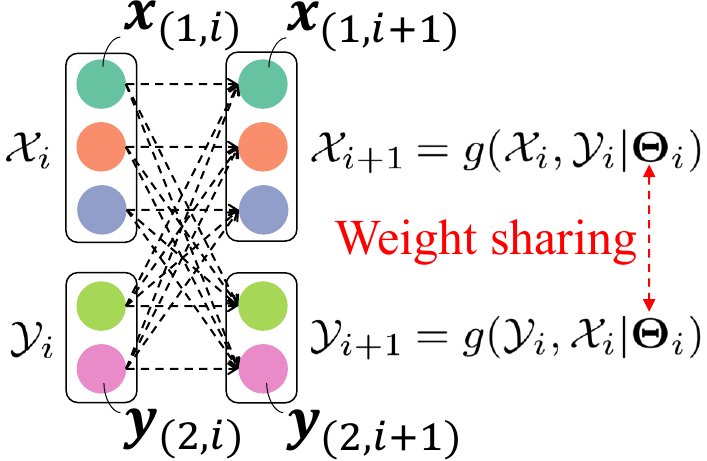}
\end{minipage}%
\begin{minipage}{.6\textwidth}
  \centering
  \includegraphics[scale = 0.30]{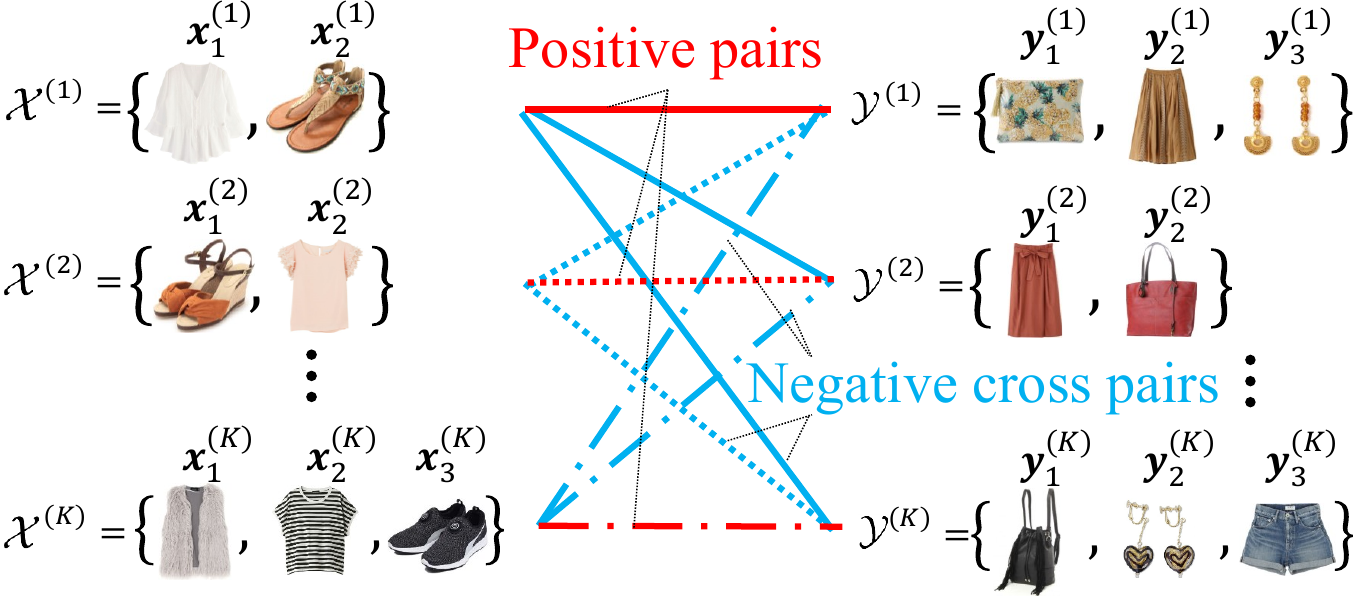}
\end{minipage}
\par
\medskip
\noindent
\begin{minipage}[t]{.37\textwidth}
  \centering
  \captionof{figure}{A diagram of CSeFT.
                   Here, we assume $|\mathcal{X}|=3$ and $|\mathcal{Y}|=2$. The colors indicate the respective set members.}
  \label{fig:crosssettrans}
\end{minipage}%
\hfill
\begin{minipage}[t]{.59\textwidth}
  \centering
  \captionof{figure}{$K$-pair-set-based matching candidates. Red and blue lines indicate correct pairs $(\mathcal{X}^{(k)}$, $\mathcal{Y}^{(k)})$ and negative cross pairs  $(\mathcal{X}^{(k)}$, $\mathcal{Y}^{(k')}): \forall k' \neq k$, where $k, k' \in \{ 1, \cdots, K\}$, respectively.}
  \label{fig:npair}
\end{minipage}
\end{figure}

Here, consider the transformation of a pair of set-feature vectors $(\mathcal{X},\mathcal{Y})$ into new feature representations on $\mathfrak{X}^N \times \mathfrak{X}^M$, using two-set-permutation equivariant functions. Let $i$ be the iteration (layer) number of the CSeFT layers. Our feature extraction then can be described as a map of
 $(\mathcal{X}_{i},\mathcal{Y}_{i}) \to (\mathcal{X}_{i+1},\mathcal{Y}_{i+1})$, where $\mathcal{X}_{i+1},\mathcal{X}_{i} \in \mathfrak{X}^N$, $\mathcal{Y}_{i+1},\mathcal{Y}_{i} \in \mathfrak{X}^M$, $\mathcal{X}_{i+1}=(\bm{x}_{(n,i+1)})_{n=1}^N$, $\mathcal{X}_{i}=(\bm{x}_{(n,i)})_{n=1}^N$, $\mathcal{Y}_{i+1}=(\bm{y}_{(m,i+1)})_{m=1}^M$, and $\mathcal{Y}_{i}=(\bm{y}_{(m,i)})_{m=1}^M$.
For example, $\bm{x}_{(n,i)} \in \mathfrak{X}$ denotes the feature vector extracted by the $i$-th layer representing the $n$-th item, $\bm{x}_{n}$, and $\bm{y}_{(m,i)}$ is defined similarly. Note that the initial feature vectors with $i=0$ are found with a typical feature extractor, i.e., a deep convolutional neural network (CNN) for the image of each item. Then, we construct a parallel architecture of CSeFT, with an asymmetric transformation $g$, as follows:
\begin{eqnarray}
\mathrm{cross\mathchar`-set \, feature \, transformation \, (CSeFT)}:
\left\{
\begin{array}{ll}
\mathcal{X}_{i+1} &= g(\mathcal{X}_{i},\mathcal{Y}_{i}|\mathbf{\Theta}_{i})\\
\mathcal{Y}_{i+1} &= g(\mathcal{Y}_{i},\mathcal{X}_{i}|\mathbf{\Theta}_{i}),
\label{eq:crosssetstrans}
\end{array}
\right.
\end{eqnarray}
where $g:\mathfrak{X}^* \times \mathfrak{X}^* \to \mathfrak{X}^*$ is a permutation equivariant function and $\mathbf{\Theta}_{i}$ is learnable weights shared in the same layer. Also, we introduce the respective residual paths~\cite{he2016deep} to Eq.~(\ref{eq:crosssetstrans}). Figure~\ref{fig:crosssettrans} shows the model of our CSeFT.

We propose two possible feature extractors for $g$: an {\it attention-based function}, and an {\it affinity-based function}. Both are constructed to assign the {\it matched} feature vectors to the {\it reference} feature vector, taking account of interactions between the two sets. For simplicity, we provide an explanation via the case of extracting the features for $\mathcal{X}$ as follows (we can easily exchange $\mathcal{X}$ and $\mathcal{Y}$ for $\mathcal{Y}$).

The attention-based function of $g$ maps $\bm{x}_{(n,i)} \to \bm{x}_{(n,i+1)}$ as follows:
\begin{equation}
\bm{x}_{(n,i+1)} = \frac{1}{|\mathcal{Y}_{i}|} \sum_{\bm{y} \in \mathcal{Y}_{i}} \left( \frac{l^{(1)}_{i}(\bm{x}_{(n,i)})^{\mathrm{T}} l^{(2)}_{i}(\bm{y})}{\sqrt{d_g}} \right)_{+} l^{(3)}_{i}(\bm{y}),
\label{eq:relu_att}
\end{equation}
where $n \in \{1, \cdots, N\}$, $\mathbf{\Theta}_i = \{ {\Theta}^{(1)}_{i}, {\Theta}^{(2)}_{i}, {\Theta}^{(3)}_{i} \}$, ${\Theta}^{(j)}_{i} \in \mathbb{R}^{d_g \times d}$, $|\mathcal{Y}_{i}|=M$, $l^{(j)}_i$ denote a linear transformation, i.e., $l^{(j)}_i(\bm{x}):=\Theta^{(j)}_{i} \bm{x}$, and $\left( \right)_+$ is a non-negative mapping, i.e., ReLU~\cite{glorot2011deep}, which introduces nonlinear interactions between the two elements.

Note that Eq.~(\ref{eq:relu_att}) relates to other attention models~\cite{jain2019attention,ilse2018attention,yang2016hierarchical,hu2018squeeze,DBLP:journals/corr/VaswaniSPUJGKP17,lee2019set}, especially the dot-product attention~\cite{DBLP:journals/corr/VaswaniSPUJGKP17,lee2019set}. The dot-product attention has been introduced to calculate the weighted average on $\mathcal{Y}$ using $\mathrm{softmax}$
as the coefficients. However, the $\mathrm{softmax}$ operation would be inconsistent with our matching objective, as through normalization it increases the coefficients even in unmatched cases of $\mathcal{X}$ and $\mathcal{Y}$. To preserve non-linearity, we use the non-negative weighted sum instead and average it.

The affinity-based function of $g$ maps $\bm{x}_{(n,i)} \to \bm{x}_{(n,i+1)}$ as follows:
\begin{equation}
\bm{x}_{(n,i+1)} = \frac{1}{2}\left( \bar{\bm{x}}_{(n,i)} + \frac{1}{|\bar{\mathcal{Y}}_{i}|} \sum_{\bar{\bm{y}} \in \bar{\mathcal{Y}}_{i}} \left( \frac{\bar{\bm{x}}_{(n,i)}^{\mathrm{T}} \bar{\bm{y}} }{\sqrt{d_g}} \right)_{+} \bar{\bm{y}} \right),
\label{eq:sim}
\end{equation}
where  $\mathbf{\Theta}_i = \{ {\Theta}^{(1)}_{i}, {\Theta}^{(2)}_{i} \}$, $\bar{\bm{x}}_{(n,i)} = l^{(1)}_{i}(\bm{x}_{(n,i)})$, and $\bar{\mathcal{Y}}_{i} = \{l^{(2)}_{i}(\bm{y}_{(m,i)})\}_{m=1}^M$. Using the two linear transformations $l^{(1)}_i$ and $l^{(2)}_i$, the affinity-based function combines the resembling feature vectors within different sets so that the feature vectors for $\mathcal{X}$ have similar representations to the linearly transformed vectors in $\mathcal{Y}$.

Other simple permutation equivariant functions of $g$, e.g., $\bm{x}_{(n,i+1)}=\bm{x}_{(n,i)} + \frac{1}{|\mathcal{Y}_i|}\sum_{\bm{y} \in \mathcal{Y}_i}\bm{y}$, may be utilized. However, we consider it a function incapable of extracting appropriate enough features without any rich interactions between the two sets to yield accurate matching for two sets.

Instead of performing $g$ singly, we introduce a multihead structure~\cite{DBLP:journals/corr/VaswaniSPUJGKP17} to our feature extractor $g$, which is also a permutation equivariant function. Denoting the output of $g(\mathcal{X}_i, \mathcal{Y}_i|\mathbf{\Theta}^{(j)}_i)$ as $g^{(j)}_{\mathcal{X}_i}$, the multihead version of $g$ is defined as $\Theta_h \mathrm{Concat} \left( g^{(1)}_{\mathcal{X}_i}, \cdots, g^{(h)}_{\mathcal{X}_i} \right)$, where $\mathrm{Concat}$ indicates a concatenation for each corresponding set member in $g^{(j)}_{\mathcal{X}_i}$, $\Theta_h \in \mathbb{R}^{d \times h {d_g}}$, and $h {d_g} = d$. Note that the multihead structure is related to recent models such as MobileNet~\cite{DBLP:journals/corr/HowardZCKWWAA17}, which isolates and places the convolutional operations in parallel to reduce the calculation costs whilst preserving the accuracy of the recognition. We assume that the multihead structure provides various interactions between the set members, reducing the calculation costs as well.

Note that we can stack CSeFTs or combine it with other networks that operate upon the sets or items independently, which preserves the symmetric architecture. Although a function of our CSeFT does not entail interactions within a set, stacking CSeFTs takes account of higher-order interactions between multiple elements involving the intra-set, which is a similar way of overlaying convolution layers for CNNs. We discuss the overall architecture in Section~\ref{subsec:arch}. 

\subsection{Calculating Matching Score for Sets} \label{subsec:as}
We introduce a matching layer to calculate the matching score between two given sets, mapping $2^\mathfrak{X}\times 2^\mathfrak{X} \to [0,\infty]$. It is designed to calculate the inner product for every combination of set members across sets, so we call this {\it cross-similarity} (CS), defined as follows:
\begin{equation}
 \mathrm{CS}(\mathcal{X}, \mathcal{Y}|W) := \frac{1}{|\mathcal{X}| |\mathcal{Y}|}\sum_{\bm{x} \in \mathcal{X}} \sum_{\bm{y} \in \mathcal{Y}} \left(\frac{l(\bm{x}|W)^\mathrm{T} l(\bm{y}|W)}{\sqrt{d_w}}\right)_+,
\label{eq_CS}
\end{equation}
where $x$ and $y$ are feature vectors in $\mathcal{X}$ and $\mathcal{Y}$, respectively, $l$ is a linear function allowing conversions into a lower-dimensional space using learnable weights $W \in \mathbb{R}^{d_w \times d}$, i.e., $l(\bm{x}|W):=W\bm{x}$, and $d_w$ is the number of dimensions of the lower-dimensional space. CS can be seen as a calculation of the average similarity in the linear subspaces created by the dimensionality reduction $l$, or the normalized and non-negative inner product if both sets contain only one set member. 

While CS function is based on a simple pair relationship, considering the feature extraction provided by CSeFT layers that involves multiple elements to represent each feature, thereby higher-order relationships are included.

Instead of calculating CS singly, we utilize multiple CSs (mCS) to combine the CSs calculated with different linear mappings. The procedure runs as follows:
\begin{equation}
\mathrm{mCS}(\mathcal{X},\mathcal{Y}|\mathbf{W}) = l(\mathrm{Concat} \left( \mathrm{CS}_1, \cdots, \mathrm{CS}_{h'} \right)|W_o),
\label{eq_mCS}
\end{equation}
where $\mathbf{W} = \{ W_1, \cdots, W_{h'}, W_o\}$, $\mathrm{CS}_j = \mathrm{CS}(\mathcal{X}, \mathcal{Y}|W_j) \in \mathbb{R}$, the linear function $l$ with learnable weights $W_o$ maps $\mathbb{R}^{h'} \to \mathbb{R}$, and $h' {d_w} = d$.

Since CS and mCS are symmetric and permutation invariant functions, combined with the fact that CSeFT is a two-set-permutation equivariant function, our model is symmetric and invariant under any permutation of items within each set in these properties.

\subsection{Training for Pairs of Sets} \label{subsec:train}

Next, the task of maximizing the matching score is translated into a minimization of a loss function of set matching, allowing for comparison against the scores for other matching candidates. Although several studies have investigated loss functions for point-to-point~\cite{NIPS2016_6200,mishchuk2017working} or point-to-set~\cite{yu2018hard,zhou2017point} metric learning, the loss function for set matching has not been well studied, and the efficient approach to preparing $K$ candidates for each query set is non-trivial.

To train our model efficiently, we create matching candidates from the correct pairs, as described in Figure~\ref{fig:npair}. Let $(\mathcal{X}^{(k)},\mathcal{Y}^{(k)})$ be a correct pair of sets, where $k \in \{1,\cdots,K\}$. From those $K$-pair, by extracting all $\mathcal{Y}^{(k)}$, we create the set of $\mathcal{Y}^{(k)}$ as $\mathbf{Y}=\{\mathcal{Y}^{(1)},\cdots,\mathcal{Y}^{(K)}\}$. That is, $\mathbf{Y}$ is composed of sets exhibiting correct relations to the respective $\mathcal{X}^{(k)}$, and $\mathbf{Y}$ can be used as a set of candidates for each $\mathcal{X}^{(k)}$ in the training stage. We construct positive pairs and negative cross pairs from these candidates by assuming that one correct pair exists for the respective sets, as described in Section~\ref{sec:Preliminaries}. Then, we train our models using these pairs with a conventional softmax cross-entropy loss. We consider the above training method as a set version of $N$-pair loss~\cite{NIPS2016_6200}, so we call this $K$-pair-set loss.

\section{Related Works} \label{sec:Related_Work}
\noindent {\bf Set-Input Neural Networks.}
Deep learning architectures for set data have been well studied~\cite{vinyals2015order,lee2019set,DBLP:journals/corr/ZaheerKRPSS17,DBLP:journals/corr/QiSMG16}, and several studies have investigated its representation universality~\cite{yarotsky2018universal,wagstaff2019limitations,sannai2019universal,segol2019universal}.
In the work of Lee et al.~\cite{lee2019set}, the state-of-the-art set-feature model, Set Transformer, was introduced by applying a self-attention based Transformer~\cite{DBLP:journals/corr/VaswaniSPUJGKP17} to a set data. Set Transformer is trained through supervised/unsupervised learning and transforms a set data into a vector/matrix representation to recognize set features.
However, constructing a deep learning model that can manage multiple sets has not been well studied.

\noindent {\bf Set-to-Set Matching.}
Various studies have suggested modeling a set as a hull~\cite{cevikalp2010face,hu2011sparse,yang2013face,zhu2013point}, hyperplanes~\cite{vincent2002k,gionis1999similarity}, linear subspace~\cite{yamaguchi1998face,kim2007discriminative,wang2008manifold,hamm2008grassmann}, convex cone~\cite{sogi2018method}, covariance matrix~\cite{wang2012covariance,cai2010matching}, Gaussian model~\cite{shakhnarovich2002face,arandjelovic2005face}, among others, for matching sets. The methods above do not include feature learning schemes for paired sets, and most of them require specific computations involving optimization methods to measure the similarity/distance between the set models. Compared with these optimization-based methods, our models are based on a feed-forward function, thus are potentially easier to scale up.

\noindent {\bf Applications.}
Many fashion item recommendation studies have investigated natural combinations of fashion items, the so-called visual fashion compatibility, to recommend fashion items or outfits~\cite{han2017learning,he2016learning,hsiao2018creating,DBLP:journals/corr/abs-1803-09196,DBLP:journals/corr/LiCZL16}. In this study, the main difficulties of the fashion set matching procedures lie in satisfying the fashion compatibility requirements of the matched sets.

In the applications of group re-id~\cite{DBLP:journals/corr/LisantiMBF17,Xiao:2018:GRL:3240508.3240539,lin2019group,zheng2009associating,cai2010matching,8919493,zhu2016consistent}, 
problems of multiple instance matching arise. One group re-id scenario has been proposed that the detection of known groups from videos~\cite{lin2019group} is required. Also, two group re-id datasets, the Road Group dataset, and the DukeMTMC Group dataset\footnote{Note that the DukeMTMC~\cite{ristani2016performance} is no longer available.} have been constructed~\cite{lin2019group}, which include bounding box annotations for each person. Our experiments focus on set-to-set matching using these given cropped images.

\noindent {\bf Methods for Non-Exchangeable Data.}
Many powerful data-processing methods, such as graph matching~\cite{bai2019simgnn,guo2018neural,li2019graph,bai2018convolutional,zanfir2018deep,fey2020deep,10.1145/3292500.3330845}, graph classification~\cite{maron2018invariant}, entity matching~\cite{mudgal2018deep}, and sequence matching~\cite{si2018dual,caspi2006feature} have been proposed based on the specific data structures. In natural language processing, Devlin et al. achieved state-of-the-art results in various tasks using the bidirectional encoder representations from transformers (BERT)~\cite{devlin2018bert}. Furthermore, Cucurull et al. applied graph neural networks (GNNs) to predict fashion compatibility between related fashion items using graph structures~\cite{cucurull2019context}. Although the data in those tasks are known to be non-exchangeable, we still consider that comparing these promising models with our model is possible and necessary.

\section{Experiments} \label{sec:Experiments}

\subsection{Overall Architecture} \label{subsec:arch}
In this section, we briefly describe our models. Borrowing from the encoder--decoder structure~\cite{DBLP:journals/corr/VaswaniSPUJGKP17,lee2019set,DBLP:journals/corr/NewellYD16}, we construct our architecture by combining the encoder~\cite{lee2019set}, called a self-attention block, with the decoder of our CSeFT. We apply a feed-forward network comprising two fully connected layers with a leaky ReLU~\cite{maas2013rectifier} to the first argument of each function $g$. We then repeat this structure $L$ times in succession, as described in Figure~\ref{fig:block}. 
We set $h$, $h'$, $L$, and $d$ to 8, 8, 2, and 512, respectively. 
To combine it with CNN features, for the fashion task, we use the Inception-v3~\cite{szegedy2016rethinking}, which is pre-trained using the ILSVRC-2012 ImageNet~\cite{russakovsky2015imagenet}, and finetune it. We extract the feature vectors on $\mathbb{R}^{2048}$ from the global average pooling layer and linearly transform it into $\mathbb{R}^{512}$. For the group re-id tasks, we utilize a simple CNN that maps $3 \to$ $64 \to$ $128 \to$ $256 \to 512$ channels using $3\times3$ kernels and train it from scratch.

\subsection{Baselines for Comparisons} \label{subsec:Baselines}

We validate our architecture through comparison with other set-matching models. However, to the best of our knowledge, studies using deep neural networks for matching two heterogeneous sets are non-existent. Instead, we use extensions from the state-of-the-art set-input method and the promising models in other related domains to a set-to-set matching procedure as described below, and consider this acceptable for the comparison. We also present ablation studies including other ordinary set matching functions.

\noindent {\bf Set Transformer.} The Set Transformer~\cite{lee2019set} transforms a set of feature vectors into a vector on $\mathbb{R}^d$. Denoting the Set Transformer model $\mathrm{ST}$, we perform the extension by calculating the matching score between the two sets $\mathcal{X}$ and $\mathcal{Y}$ via the inner product $\mathrm{ST}(\mathcal{X})^\mathrm{T} \mathrm{ST}(\mathcal{Y})$, sharing the weights between the two $\mathrm{ST}$.

\noindent {\bf BERT.} We consider a union of two sets as a set-input for the extension of BERT~\cite{devlin2018bert} and omit the individual token embedding, i.e., the position embedding. We use the segment embedding to designate items of $\mathcal{X}$ and $\mathcal{Y}$. We use three variants: BERT$_\mathrm{BASE}$ is the same model as described in~\cite{devlin2018bert}; BERT$_\mathrm{BASE-AP}$ uses the average pooling in the last layer; and BERT$_\mathrm{SMALL}$ is a four-layered version of BERT$_\mathrm{BASE}$ with eight heads, and the hidden size is 512.

\noindent {\bf GNN.} We combine two sets as one input for the extension of GNN~\cite{cucurull2019context}.
Because this model is not presented to train in an end-to-end with the feature extractor, we do not finetune the CNN in fashion set matching, where pre-trained CNNs are used, but train it in an end-to-end manner for the group re-id task. Note that we omit the context provided from the external graphs in the evaluation stage to apply this model in the same scenarios of our tasks. We set the training epoch to 256 in the group re-id to enhance the training results of the GNN.

The properties in the above models are different from ours. The extension of Set Transformer satisfies the exchangeability criteria; however, no interactions between paired sets are provided. Both extensions of BERT and GNN provide the interactions but do not facilitate the exchangeability of two sets. 

Additionally, in our first experiments, we introduce a conventional CNN, trained by Hard-Aware Point-to-Set loss (HAP2S)~\cite{yu2018hard} as a minimum configuration. We use the exponential weighting and the same parameter setting described in \cite{yu2018hard}. We also use the {\it batch all} strategy~\cite{DBLP:journals/corr/HermansBL17} to train the CNN effectively.

\subsection{Training Settings}

In this section, we briefly describe the training settings. 
We use a stochastic gradient descent method with a learning rate of 0.005, a momentum of 0.5, and a weight decay of 0.00004.
We set the numbers of matching candidates and the training epochs to 4 and 32, 16 and 128, and 81 and 3500, for the tasks of fashion set matching, Market-1501 Group dataset, and Road Group dataset, respectively. 
We train both the CNN and set-matching model in an end-to-end manner. In each iteration, we randomly swap pairs of sets and items in each set, and randomly flip images horizontally, to learn all the methods stably.

\subsection{Fashion Set Matching}

\noindent {\bf Dataset.}
We examine the set matching task for fashion recommendation, using the {\it IQON dataset}~\cite{DBLP:journals/corr/abs-1807-03133}. 
The dataset consists of recently created, high-quality outfits, including 199,792 items grouped into 88,674 outfits. We split these outfits into groups, using 70,997 for training, 8,842 for validation, and 8,835 for testing. 

Our task can be considered an extended version of a standard task, Fill-In-The-Blank~\cite{cucurull2019context}, which requires us to select an item that best extends an outfit from among four candidates. Because selecting a set corresponds to filling multiple blanks, we consider the set matching problem as Fill-In-The-$N$-Blank.

\noindent {\bf Preparing Set Pairs.}
To construct the correct pair of sets to be matched, we randomly halve the given outfit $\mathcal{O}$ into two non-empty proper subsets $\mathcal{X}$ and $\mathcal{Y}$ as follows: $\mathcal{O} \to \{\mathcal{X}, \mathcal{Y}\}$, where $\mathcal{X} \cap \mathcal{Y} = \emptyset$. Here, we extend this setting to include more general situations. We select $Q$ outfits $\{\mathcal{O}^{(1)}, \cdots, \mathcal{O}^{(Q)}\}$ randomly and split the respective outfits in half $\mathcal{O}^{(q)} \to \{\mathcal{X}^{(q)}, \mathcal{Y}^{(q)}\}$, where $q \in \{1,\cdots,Q\}$. We regard the two sets $\{\mathcal{X}^{(1)},\cdots,\mathcal{X}^{(Q)}\}$ and $\{\mathcal{Y}^{(1)},\cdots,\mathcal{Y}^{(Q)}\}$ as the correct pair, which consists of $Q$ fashion styles. In the training phase, we set $Q = 4$.

\noindent {\bf Fashion Set Matching.}
We discuss the experimental results of the fashion set matching. 
Table~\ref{tab:outfit} shows significantly different results between our models and the baselines. Here, Cross Attention and Cross Affinity denote our models with the attention-based and affinity-based functions, respectively. 
Comparing the performance of Cross Affinity and BERT$_\mathrm{SMALL}$, which is the most accurate among the baselines, the differences in their accuracy were 5.2\%, on average, 8.9\%, at maximum, where the differences were relatively significant in the complicated setting on Mix:4. Furthermore, Table~\ref{tab:outfit} shows that the affinity-based function performed better than the attention-based one.

In this experiment, we consider that the components on which the comparative effectiveness of the proposed models depended were potentially three-fold. Compared with the extensions of BERT, (a) our model preserves the exchangeability in two sets, which may ensure that the set features to be matched are accurately represented. Furthermore, (b) our model preserves two set features explicitly, whereas BERT provides a set of features with segment embedding that may have a limitation. Compared with the results of the Set Transformer, our models and BERT yielded accurate results is made possible by (c) providing the strength of interactions between two sets. Therefore, we conclude that these results justify the fine aspects of our architecture.

\begin{table}[t]
\centering
\caption{Accuracy of fashion set matching (\%). Cand and Mix indicate the number of matching candidates and number of mixed outfits $(Q)$, respectively.}
\resizebox{0.6\columnwidth}{!}{
\begin{tabular}{@{}lrrrrrr@{}}
\toprule
 & \multicolumn{3}{c|}{Cand:4} & \multicolumn{3}{c}{Cand:8} \\
\multicolumn{1}{c}{Method} & \multicolumn{1}{l}{Mix:1} & Mix:2 & \multicolumn{1}{c|}{Mix:4} & \multicolumn{1}{l}{Mix:1} & Mix:2 & Mix:4 \\ \midrule
Set Transformer & 68.0 & 73.5 & 65.3 & 50.5 & 57.5 & 49.6 \\
BERT$_\mathrm{SMALL}$ & 82.1 & 87.3 & 69.7 & 69.9 & 77.0 & 53.0 \\
BERT$_\mathrm{BASE}$ & 81.4 & 86.6 & 66.1 & 69.2 & 76.3 & 50.8 \\
BERT$_\mathrm{BASE-AP}$ & 80.8 & 86.4 & 65.4 & 68.6 & 75.7 & 49.5 \\
GNN & 35.4 & 32.4 & 25.5 & 19.9 & 17.5 & 13.4 \\
HAP2S & 39.3 & 36.6 & 32.0 & 23.3 & 20.8 & 17.8 \\
Cross Attention (ours) & 80.8 & 88.8 & 74.3 & 68.9 & 80.6 & 58.9 \\
Cross Affinity (ours) & \textbf{85.1} & \textbf{90.6} & \textbf{75.9} & \textbf{73.8} & \textbf{82.8} & \textbf{61.9} \\ \bottomrule
\end{tabular}
}
\label{tab:outfit}
\end{table}

\subsection{Group Re-Identification}

We present the results of a group re-id on the Market-1501 Group dataset, a new extension of a well-known person re-id dataset, Market-1501~\cite{zheng2015scalable}, and the Road Group dataset~\cite{Xiao:2018:GRL:3240508.3240539}. The task is to identify the pairs of sets that consist of individual images of the (mostly) same multiple persons under noisy situations.

\noindent {\bf Evaluation on Market-1501 Group Dataset.} 
We constructed the training/validation data based on query/gallery splits provided in~\cite{zheng2015scalable}. Because there are few person images provided for each camera position, we did not consider camera information.
We regard sets of gallery and query data as $\mathcal{X}$ and $\mathcal{Y}$, respectively, where each set contains three persons in non-noisy cases, and each person is represented by three different images.

We investigated noise robustness through the experiments to show that our models do not over-fit on the data; here, the {\it noise} means that random persons that accidentally contained into the group additionally or that the label noise~\cite{DBLP:journals/corr/abs-1712-05055} for paired sets generated based on the given noise fraction. Note that the noise persons and label noise have some relations, e.g., a candidate set composed of only noise persons corresponds to a set mislabelled by label noise.

Table~\ref{tab:reid} presents the comparison results.
In the non-noisy case, many models showed almost perfect accuracies; we consider that {\it averaging feature vectors in sets} achieves high accuracy in this homogeneous case.
In the case the noise person included, the noise ratio was inversely proportional to the accuracy across all the models; however, our models yielded more accurate results, e.g., the average accuracy of Cross Affinity, Set Transformer, and BERT$_\mathrm{BASE-AP}$ was 87.0, 80.6, and 72.4\%, respectively. Because the main differences between the architectures exist in the interactions for paired sets or the exchangeability, the results support the claim that considering these properties improves both the accuracy and robustness. Furthermore, in the case of label noise fraction is $0.8$, the permutation invariance would be essential to preserve high accuracy.

\noindent {\bf Evaluation on Road Group Dataset.} 
We conduct experiments on the Road Group dataset~\cite{Xiao:2018:GRL:3240508.3240539,lin2019group}, which consists of 162 group pairs taken from a 2-camera-view of a crowded road scene. One image per group for each camera is provided, where most groups do not have the same person's image in common with the different group pairs.
Following the experimental protocol described in~\cite{Xiao:2018:GRL:3240508.3240539,lin2019group}, we construct training/validation datasets, splitting the 162 group pairs randomly in half into two different 81 group pairs, and reporting the accuracies calculated by the cumulative matching characteristic (CMC) metric~\cite{moon2001computational}. 

Because group re-id is a newly emerging task, most datasets, including the Road Group dataset, contain a small number of groups and images, and training on such datasets is difficult~\cite{huang2019group}. Specifically, our set-to-set matching method extracts features that rely on input set pairs, thus, the variations in the set pairs are crucial. Considering the difference in appearances or camera parameters, however, importing external data~\cite{huang2019group,huang2019dot} is also a challenging task itself.

To relax the data limitation, we introduce our novel set-data augmentation (set-aug) method that significantly enhances the learning results of the proposed set-to-set matching modules by increasing the training data. Given positive person image pairs and several negative person images, creating set pairs randomly on each training iteration, our set-aug effectively increases the group member variations (please refer to Appendix~\ref{a:set-aug} for details).

\begin{table}
\centering
\caption{Accuracy (\%) for Market-1501 Group. The number of candidates is 5.}
\resizebox{0.9\columnwidth}{!}{
\label{tab:reid}
\begin{tabular}{lrrrrrrrrrr}
\toprule
 & \multicolumn{1}{c|}{Non-noisy} & \multicolumn{6}{c|}{Ratio of {\it noise} persons in $\mathcal{X} \times \mathcal{Y}$} & \multicolumn{3}{c}{Label noise frac.} \\
\multicolumn{1}{c}{Method} & \multicolumn{1}{l|}{}
 & $(\frac{0}{3},\frac{1}{4})$ 
 & $(\frac{1}{4},\frac{1}{4})$ 
 & $(\frac{0}{3},\frac{3}{6})$ 
 & $(\frac{0}{3},\frac{5}{8})$ 
 & $(\frac{3}{6},\frac{3}{6})$ 
 & \multicolumn{1}{c|}{$(\frac{5}{8},\frac{5}{8})$} & 0.2 & 0.4 & 0.8 \\ 
\midrule
Set Transformer & 99.5 & 95.1 & 89.9 & 85.7 & 80.4 & 65.7 & 48.1 & \textbf{99.3} & 98.8 & 95.6 \\
BERT$_\mathrm{SMALL}$ & 94.3 & 77.6 & 69.2 & 83.7 & 64.9 & 49.5 & 24.7 & 99.2 & 98.7 & 79.5 \\
BERT$_\mathrm{BASE}$ & 96.8 & 80.5 & 77.6 & 68.8 & 69.9 & 61.9 & 49.2 & 98.9 & 98.1 & 76.0 \\
BERT$_\mathrm{BASE-AP}$ & 97.3 & 84.4 & 74.7 & 70.7 & 69.3 & 62.8 & 47.7 & \textbf{99.3} & 97.5 & 77.9 \\
GNN & 82.0 & 29.3 & 46.0 & 23.7 & 22.1 & 29.3 & 21.1 & 81.7 & 73.0 & 76.7 \\
Cross Attention (ours) & 99.6 & \textbf{96.9} & \textbf{94.8} & 91.9 & 90.7 & \textbf{72.9}  & 56.1 & \textbf{99.3} & 99.6 & 95.5 \\
Cross Affinity (ours) & \textbf{99.7} & 96.5 & 92.5 & \textbf{94.4}  & \textbf{92.4}  & 72.0 & \textbf{61.7}  & \textbf{99.3} & \textbf{99.9} & \textbf{98.4} \\
\bottomrule
\end{tabular}
}
\end{table}

\begin{table}[t]
\centering
\caption{Evaluation results (\%) for Road Group dataset. }
\resizebox{0.8\columnwidth}{!}{
\begin{tabular}{@{}llllll@{}}
\toprule
\multicolumn{1}{c}{Method (detector-based)} & \multicolumn{1}{c}{CMC-1} & \multicolumn{1}{c}{CMC-5} & \multicolumn{1}{c}{CMC-10} & \multicolumn{1}{c}{CMC-15} & \multicolumn{1}{c}{CMC-20} \\ \midrule
\multicolumn{6}{c}{\textit{Data augmentation ablation}} \\
Cross Affinity (our baseline) & 45.2 $\pm$ 3.5 & 77.5 $\pm$ 2.9 & 87.9 $\pm$ 3.8 & 91.9 $\pm$ 2.4 & 94.1 $\pm$ 2.1 \\
Baseline + img-aug & 47.7 $\pm$ 4.2 & 78.3 $\pm$ 3.2 & 87.7 $\pm$ 2.6 & 91.1 $\pm$ 2.4 & 93.3 $\pm$ 1.8 \\
Baseline + set-aug & {\bf 84.0} $\pm$ 3.6 & 93.8 $\pm$ 0.8 & {\bf 96.8} $\pm$ 0.6 & {\bf 97.0} $\pm$ 1.0 & 97.5 $\pm$ 1.1 \\
Baseline + set-aug + img-aug & 81.7 $\pm$ 1.9 & {\bf 94.1} $\pm$ 1.3 & 96.5 $\pm$ 1.1 & {\bf 97.0} $\pm$ 0.9 & {\bf 97.8} $\pm$ 0.8 \\ \midrule
Baseline + set-aug (ours) & {\bf 84.0} $\pm$ 3.6 & 93.8 $\pm$ 0.8 & {\bf 96.8} $\pm$ 0.6 & 97.0 $\pm$ 1.0 & 97.5 $\pm$ 1.1 \\
MGM w/ spatial layout~\cite{lin2019group} & 80.2 & 93.8 & 96.3 & {\bf 97.5} & 97.5 \\
MGM w/o spatial layout~\cite{lin2019group} & 70.4 & 90.1 & 91.3 & 92.6 & 96.3 \\
TSCN w/ external data ~\cite{huang2019group} & {\bf 84.0} & {\bf 95.1} & 96.3 & \multicolumn{1}{c}{-} & {\bf 98.8} \\
GNN w/ external data ~\cite{huang2019dot} & 74.1 & 90.1 & 92.6 & \multicolumn{1}{c}{-} & {\bf 98.8} \\
\toprule
\multicolumn{1}{c}{Method (GT-based)} & \multicolumn{1}{c}{CMC-1} & \multicolumn{1}{c}{CMC-5} & \multicolumn{1}{c}{CMC-10} & \multicolumn{1}{c}{CMC-15} & \multicolumn{1}{c}{CMC-20} \\ \midrule
Baseline + set-aug (ours) & {\bf 85.7} $\pm$ 3.7 & {\bf 96.3} $\pm$ 0.8 & {\bf 97.8} $\pm$ 0.5 & {\bf 98.3} $\pm$ 0.6 & {\bf 98.3} $\pm$ 0.6 \\
MGM w/ spatial layout~\cite{lin2019group} & 82.4 & 95.1 & 96.3 & 97.5 & 98.0 \\
\bottomrule
\end{tabular}
}
\label{tab:reid_road}
\end{table}

Table~\ref{tab:reid_road} shows the experimental results. The top block in Table~\ref{tab:reid_road} indicates the results of our methods with three types of data augmentation: (a) the horizontal flipping~\cite{krizhevsky2012imagenet}, which is used to train the baseline model; (b) image-based data augmentation (img-aug), which includes both scale augmentation~\cite{simonyan2014very,he2016deep} and random erasing~\cite{zhong2017random} on images; and (c) our set-aug. Using the 81 pre-defined groups, the baseline model was not very effective, even with img-aug. However, using the set-aug, our method exhibited significant improvements without applying img-aug. These results imply that generating combinations on sets is very beneficial to our models. The other parts in Table~\ref{tab:reid_road} show that our methods yield very competitive results, compared with the state-of-the-art methods that utilize a large transferred external dataset or auxiliary features such as spatial layout information within each group. Furthermore, compared with MGM w/o spatial layout~\cite{lin2019group}, which also does not use the spatial layout information, our methods significantly improved the accuracy of CMC-1 by 13.6\%.

\subsection{Ablation Study}

In this section, we report the results of an ablation study performed to highlight the importance of each proposed component. The top part in Table~\ref{tab:ablation} shows the two results obtained when our models are trained using triplet loss with the soft-margin and batch all strategy~\cite{DBLP:journals/corr/HermansBL17} or the proposed $K$-pair-set loss. Triplet loss triggered significant accuracy degradation, even though the losses converged to near zero in the training stages. On the other hand, the proposed $K$-pair-set loss can manage to accurately train the models by considering the loss of selecting paired sets among multiple candidates. The second topmost part in Table~\ref{tab:ablation} shows the results of ablations in the feature extractor. Reducing the number of layers and number of multiheads in the CSeFT, and excluding the encoder and CSeFT, the accuracies are degraded by 1.0, 1.1, 1.3, and 4.2\%. 
Our model without the encoder performed well (1.3\% degradation) but showed a somewhat slow convergence. However, excluding the CSeFT module significantly reduced the accuracy
\begin{wraptable}[27]{r}{4.7cm}
\center
\caption{Ablation study. Average accuracies (\%) on Market-1501 Group are shown, where the non-noisy and six noise person patterns, presented in Table~\ref{tab:reid}, are included.}
\resizebox{.38\columnwidth}{!}{
\begin{tabular}{@{}lc@{}}
\toprule
\multicolumn{1}{c}{Method} & Accuracy \\ \midrule
\multicolumn{2}{c}{\textit{Training method ablation}} \\
{\bf Cross Affinity (baseline)} & {\bf 87.0} \\
Baseline with triplet loss & 45.5 \\ \midrule
\multicolumn{2}{c}{\textit{Feature extractor ablation}} \\
Baseline with $L$=1 & 86.0 \\
Baseline with $h$=1 & 85.9 \\
w/o Enc & 85.7 \\ 
w/o CSeFT & 82.8 \\
\midrule
\multicolumn{2}{c}{\textit{Matching layer ablation}} \\
Single CS & 86.0 \\
w/o ReLU in mCS & 85.0 \\
Max pooling & 86.1 \\
Average pooling & 85.8 \\
Projection metric & 67.0 \\
Covariance matrix & 61.1 \\
Set kernel & 53.3 \\
Cosine similarity metric & 53.1 \\
\midrule
\multicolumn{2}{c}{\textit{Feature {\rm \&} matching layer ablation}} \\
Set Transformer & 80.6 \\
\bottomrule
\end{tabular}
}
\label{tab:ablation}
\end{wraptable}
(4.2\% degradation).
These results imply that the proposed CSeFT module is a crucial part of the set-to-set matching model architecture. The second lower part of Table~\ref{tab:ablation} shows the results of ablation study performed on the matching layer. Reducing mCS to a single CS and excluding ReLU from the CS functions reduced the accuracies of both models by 1.0 and 2.0\%, respectively. It is interesting to observe that the ReLU was more important than the number of CS functions; this demonstrated the importance of nonlinearity in the matching layer. Furthermore, replacing our mCS with max pooling, average pooling, projection metric~\cite{huang2018building}, covariance matrix~\cite{wang2012covariance,cai2010matching}, set kernel~\cite{kim2019practical}, and cosine similarity metric~\cite{nguyen2010cosine} all resulted in significant accuracy degradation implying the effectiveness of our mCS functions (see Appendix~\ref{a:model} for details). The lowermost part of Table~\ref{tab:ablation} shows the results of ablation study performed on the feature extractor and matching layer. The extension of Set Transformer, which does not include the proposed CSeFT module and CS function, yielded significant accuracy degradation.
These results show the validity of our architecture for heterogeneous set-to-set matching.

\section{Conclusion} \label{sec:Conclusion}
In this study, we investigated the heterogeneous set-to-set matching problem. We proposed a novel architecture comprising the (1) cross-set feature transformation (CSeFT) module and (2) cross-similarity (CS) function, in addition to a loss function and set-data augmentation for performing set-to-set matching. 

We showed that our architecture preserves the two types of exchangeability for a pair of sets and also the items within them, thereby satisfying the requirements of set-to-set matching procedure.

We demonstrated that our models performed well compared with the state-of-the-art methods and baselines in the fashion set matching and group re-id experiments. Furthermore, we validated our proposed architecture through the ablation study. These results support the claim that the exchangeability and the feature representations extracted with interactions between the two sets improve the accuracy and robustness of the heterogeneous set-to-set matching.

\clearpage
\bibliographystyle{splncs04}
\bibliography{egbib.bib}

\clearpage
\appendix
\appendixpage
\addappheadtotoc

\setcounter{table}{0}
\renewcommand{\thetable}{A\arabic{table}}

\section{More Details of Models} \label{a:model}

In ablation study, we replace our mCS with max pooling, average pooling, projection metric~\cite{huang2018building}, covariance matrix~\cite{wang2012covariance,cai2010matching}, set kernel~\cite{kim2019practical}, and cosine similarity metric~\cite{nguyen2010cosine}. For projection metric, we use the inner product as described in~\cite{huang2018building}. For covariance matrix, we calculate two covariance matrices and the inner product between the two matrices~\cite{zhu2013point}. Note that we also normalize the calculated similarities as described in~\cite{nguyen2010cosine}. For set kernel, we use Gaussian kernel and multiple kernel
learning as described in~\cite{li2017mmd}. 

\section{Set-Data Augmentation} \label{a:set-aug}
In this appendix, we describe our set-data augmentation (set-aug) method. Algorithm~\ref{alg:setaug} shows the set-aug algorithm. As described in the paper, given positive person image pairs $X$ and several negative person images $Z$, we create set pairs randomly on each training iteration. Here, index $i$ is the iteration number in each epoch.

We use the set-aug on Road Group dataset using Algorithm \ref{alg:setaug}. In each training iteration, we choose the number of base-set-size $s \in \{3,4\}$ randomly, and select $s-1$ paired images using $randomSelectPairedImage$. Furthermore, we add one noise image to each set randomly with a probability of 85\%, and drop an image from each set randomly with a probability of 50\%.

\begin{algorithm}[tbhp]
\caption{Set-Data Augmentation.}
\label{alg:setaug}
\setstretch{1.2}
\DontPrintSemicolon
\nl\KwData{paired-image dataset $X$, noise-image dataset $Z$, index $i$}
\nl\KwResult{paired sets $(\mathcal{X}, \mathcal{Y})$}
\nl\Begin{
  \nl//select an image-pair and create initial paired sets from $i$-th paired-image in $X$, where $|\mathcal{X}|=|\mathcal{Y}|=1$ \;
  \nl$(\mathcal{X}, \mathcal{Y}) \longleftarrow selectPairedImage(X, i)$ \;
  \nl//randomly select multiple paired images \;
  \nl$(\mathcal{X'}, \mathcal{Y'}) \longleftarrow randomSelectPairedImage(X)$ \;
  \nl$\mathcal{X} \longleftarrow \mathcal{X} \cup \mathcal{X'}$ \;
  \nl$\mathcal{Y} \longleftarrow \mathcal{Y} \cup \mathcal{Y'}$ \;  
  \nl//randomly drop the image(s) and use the remained set\;
  \nl$\mathcal{X} \longleftarrow randomDrop(\mathcal{X})$ \;
  \nl$\mathcal{Y} \longleftarrow randomDrop(\mathcal{Y})$ \;
  \nl//randomly select the noise image(s) (if possible, select the images captured on the same camera of each target set)\;
  \nl$\mathcal{X''} \longleftarrow randomSelectImage(Z)$ \;
  \nl$\mathcal{Y''} \longleftarrow randomSelectImage(Z)$ \;
  \nl$\mathcal{X} \longleftarrow \mathcal{X} \cup \mathcal{X''}$ \;
  \nl$\mathcal{Y} \longleftarrow \mathcal{Y} \cup \mathcal{Y''}$ \;  
}
\end{algorithm}

\section{More Details of IQON Dataset}
IQON (www.iqon.jp) is a user-participating fashion web service sharing outfits for women. IQON Dataset~\cite{DBLP:journals/corr/abs-1807-03133} contains images with $480\times480$ size.

To create our training dataset from IQON dataset, we set the maximum and minimum numbers of items for each outfit as eight and four, respectively; if the outfit contains more than eight items, then we randomly select eight items from it. The outfits contain roughly 5.5 items on average. After this operation, we created our training datasets.

\section{Additional Experiments}

\subsection{Subset Matching for Fashion Set Recommendation}
In this appendix, we consider a different variation of the fashion set matching task to include item category restrictions and focus on the case of $Q = 1$, where $Q$ is the number of outfits mixed in the set. We call this task subset matching.

In subset matching, for evaluation, $K$ subsets $\{\mathcal{Y}^{(1)},\cdots,\mathcal{Y}^{(K)}\}$ are provided as a set of matching candidates to the reference subset $\mathcal{X}$, while maintaining the category restrictions for each fashion item. That is, these $K$ candidates only contain same-category fashion items, e.g, tops and bottoms. 

For training, we also give the item category restriction. Note that without any category restrictions, the models tend to be trained to select the candidate $\mathcal{Y}^{(k)}$ that contains non-overlapped fashion category items, e.g., shoes, with $\mathcal{X}$ when we train the model in the case of $Q = 1$. To avoid this situation, we introduce the category restrictions to the $K$ candidates in training/testing phases.

To implement the item category constraint described above, we use the triplet loss with softplus function~\cite{DBLP:journals/corr/HermansBL17} in the subset matching problem. Here, we prepare a negative set $\mathcal{Y}^{(n)}$ by selecting random items under the category restrictions. Then, we train the models using the reference set $\mathcal{X}$ as an anchor set and the subset $\mathcal{Y}^{(p)}$ as a positive set, where $\mathcal{X}\cup\mathcal{Y}^{(p)}$ and $\mathcal{X}\cup\mathcal{Y}^{(n)}$ corresponds to the given complete outfits and unmatched outfits, respectively.

Table~\ref{tab:outfit2} shows the comparison results. Our models showed significant improvements compared with baseline models.

\begin{table}[tbh]
\centering
\caption{Accuracy of subset matching (\%). Cand indicates the number of candidates to be matched.}
\begin{tabular}{@{}llc@{}}
\toprule
\multicolumn{1}{c}{Method} & \multicolumn{1}{c}{Cand:4} & Cand:8 \\ \midrule
Set Transformer & 39.2 & 22.7 \\
BERT$_\mathrm{SMALL}$ & 50.5 & 33.8 \\
BERT$_\mathrm{BASE}$ & 50.5 & 33.5 \\
BERT$_\mathrm{BASE-AP}$ & 50.0 & 33.5 \\
GNN & 30.3 & 17.3 \\
HAP2S & 29.4 & 16.8 \\
Cross Attention (ours) & 58.1 & 41.9 \\
Cross Affinity (ours) & \textbf{60.2} & \textbf{43.3} \\ \bottomrule
\end{tabular}
\label{tab:outfit2}
\end{table}

\subsection{Weak Point Analysis} \label{wpa}
\begin{figure}[tbh]
\centering
\includegraphics[width=0.9\textwidth]{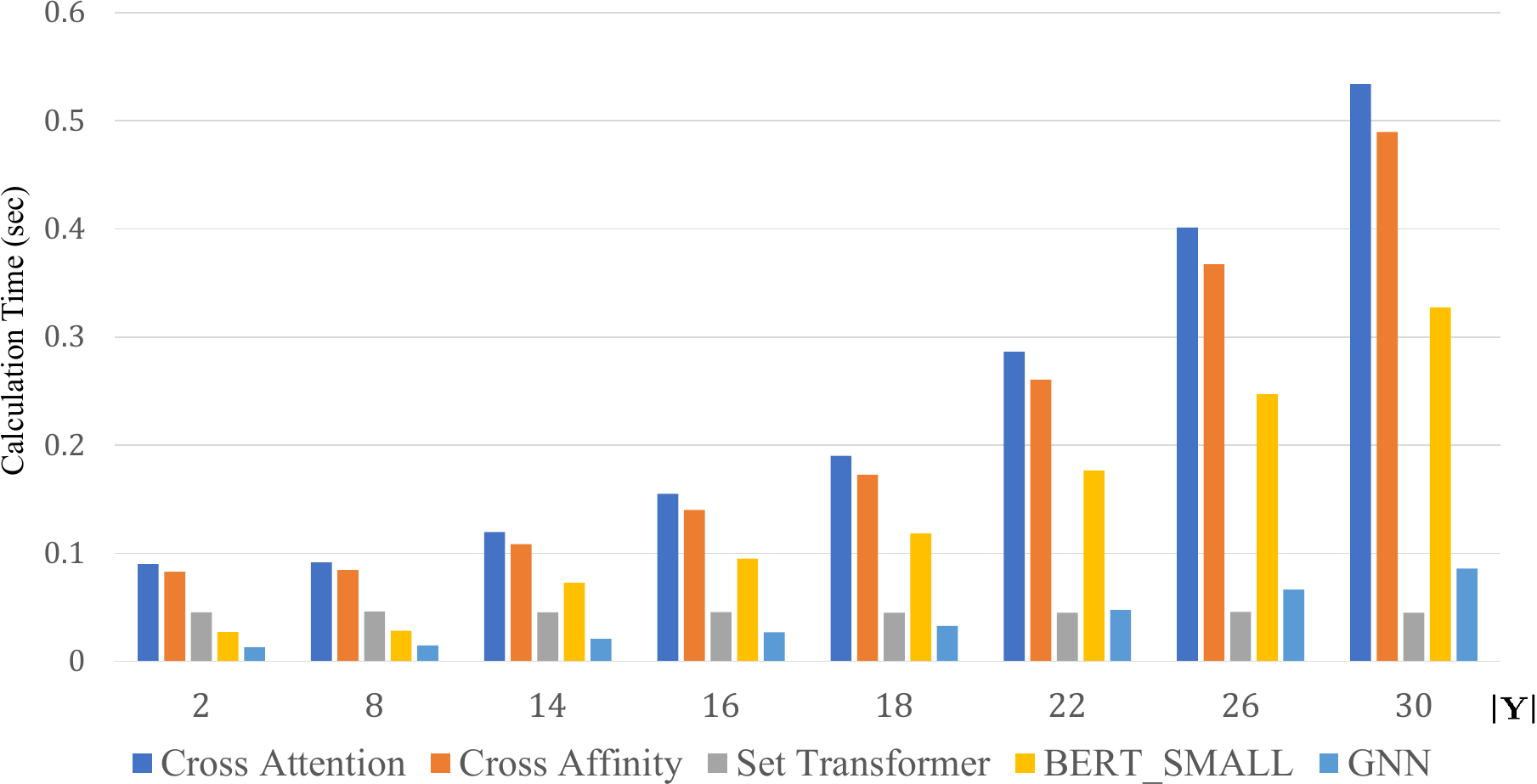} 
\caption{Inference time for set-to-set matching. Here, we test each model 110 times successively and plot the median in the last 100 records. We randomly generated pseudo data for the calculation, which are sets of vectors on $\mathbb{R}^{512}$. Each set contains eight data. We used GeForce GTX 970 for the calculation. }
\label{fig:time}
\end{figure}

The main weak point of our models is in calculation cost. We consider that our models are promising to match a reference and candidate sets in high accuracy, but impose more substantial calculations. For example, a one-set-input function, i.e., the extension of Set Transformer, can transform a set of features individually for two sets to match. Also, after the feature extractions, it does not require calculations except the inner product in matching two vectors. Comparing with the Set Transformer, our models and the extensions of BERT and GNN models need additional calculation costs in matching two sets; they need paired sets for the feature extraction. Figure~\ref{fig:time} shows the calculation time in the testing stage, where $\bf{|Y|}$ indicates the number of candidate sets. The calculation time of these models except for the Set Transformer significantly increased when the number of candidate sets increased.

Reducing the calculation costs preserving the interactions is challenging but interesting, and we leave it as future work.

\section{Limitations}
In this appendix, we discuss several issues or ideas of our models and consider them as future works.

\noindent {\bf Imbalanced Samples.} Our $K$-pair-set loss may suffer from the imbalance between positive and negative training samples when $K$ is large, leading to bad performance; thus, a strategy of hard sample mining~\cite{mishchuk2017working} may be needed.

\noindent {\bf Simple Interactions.} Our simple CSeFT modules only include interactions from inter-sets. Introducing interactions between intra-sets and inter-sets into the CSeFT module may improve set matching results.

\noindent {\bf Matching Per Paired Sets.} As described in Appendix~\ref{wpa}, one of the limitations of the set-to-set matching model is the computation cost. Currently, it is not easy to apply to larger-scale search/retrieval-like tasks.

\noindent {\bf Feature Representation.} We consider that introducing regularization terms will improve matching results via mapping the same person's feature vectors into similar or identical representations in group re-id tasks.

\end{document}